# REGAL: A Regularization based Algorithm for Reinforcement Learning in Weakly Communicating MDPs

**Peter L. Bartlett**
Computer Science Division, and
Department of Statistics
University of California at Berkeley
Berkeley, CA 94720

**Ambuj Tewari**
Toyota Technological Institute at Chicago
6045 S. Kenwood Ave.
Chicago, IL 60615

**Abstract**

We provide an algorithm that achieves the optimal regret rate in an unknown weakly communicating Markov Decision Process (MDP). The algorithm proceeds in episodes where, in each episode, it picks a policy using regularization based on the *span* of the optimal bias vector. For an MDP with $S$ states and $A$ actions whose optimal bias vector has span bounded by $H$, we show a regret bound of $\tilde{O}(HS\sqrt{AT})$. We also relate the span to various diameter-like quantities associated with the MDP, demonstrating how our results improve on previous regret bounds.

## 1 INTRODUCTION

In reinforcement learning, an agent interacts with an environment while trying to maximize the total reward it accumulates. Markov Decision Processes (MDPs) are the most commonly used model for the environment. To every MDP is associated a state space $\mathcal{S}$ and an action space $\mathcal{A}$. Suppose there are $S$ states and $A$ actions. The parameters of the MDP then consist of $S \cdot A$ state transition distributions $P_{s,a}$ and $S \cdot A$ rewards $r(s, a)$. When the agent takes action $a$ in state $s$, it receives reward $r(s, a)$ and the probability that it moves to state $s'$ is $P_{s,a}(s')$. The agent does not know the parameters $P_{s,a}$ and $r(s, a)$ of the MDP in advance but has to learn them by directly interacting with the environment. In doing so, it faces the *exploration vs. exploitation* trade-off that Kearns and Singh [2002] succinctly describe as,

> "... should the agent exploit its cumulative experience so far, by executing the action that currently seems best, or should it execute a different action, with the hope of gaining information or experience that could lead to higher future payoffs? Too little exploration can prevent the agent from ever converging to the optimal behavior, while too much exploration can prevent the agent from gaining near-optimal payoff in a timely fashion."

Suppose the agent uses an algorithm $\mathcal{G}$ to choose its actions based on the history of its interactions with the MDP starting from some initial state $s_1$. Denoting the (random) reward obtained at time $t$ by $r_t$, the algorithm's expected reward until time $T$ is

$$R^{\mathcal{G}}(s_1, T) = \mathbb{E}\left[\sum_{t=1}^{T} r_t\right] .$$

Suppose $\lambda^\star$ is the optimal per-step reward. An important quantity used to measure how well $\mathcal{G}$ is handling the exploration vs. exploitation trade-off is the *regret*,

$$\Delta^{\mathcal{G}}(s_1, T) = \lambda^\star T - R^{\mathcal{G}}(s_1, T) .$$

If $\Delta^{\mathcal{G}}(s_1, T)$ is $o(T)$ then $\mathcal{G}$ is indeed learning something useful about the MDP since its expected average reward then converges to the optimal value $\lambda^\star$ (which can only be computed with the knowledge of the MDP parameters) in the limit $T \to \infty$.

However, asymptotic results are of limited value and therefore it is desirable to have *finite time* bounds on the regret. To obtain such results, one has to work with a restricted class of MDPs. In the theory of MDPs, four fundamental subclasses have been studied. Unless otherwise specified, by a *policy*, we mean a deterministic stationary policy, i.e. simply a map $\pi : \mathcal{S} \to \mathcal{A}$.

**Ergodic** Every policy induces a single recurrent class, i.e. it is possible to reach any state from any other state.

**Unichain** Every policy induces a single recurrent class plus a possibly empty set of transient states.

**Communicating** For every $s_1, s_2 \in \mathcal{S}$, there is *some* policy that takes one from $s_1$ to $s_2$.

**Weakly Communicating** The state space $\mathcal{S}$ decomposes into two sets: in the first, each state is reachable from every other state in the set under some policy; in the second, all states are transient under all policies.

Unless we modify our criterion, it is clear that regret guarantees cannot be given for general MDPs since the agent might be placed in a state from which reaching the optimally rewarding states is impossible. So, we consider the most general subclass: *weakly communicating* MDPs. For such MDPs, the optimal gain $\lambda^\star$ is state independent and is obtained by solving the following *optimality equations*,

$$h^\star + \lambda^\star \mathbf{e} = \max_{a \in \mathcal{A}} \left( r(s,a) + P_{s,a}^\top h^\star \right) . \quad (1)$$

Here, $h^\star$ is known as the *bias* vector. Note that if $h^\star$ is a bias vector then so is $h^\star + c\mathbf{e}$ where $\mathbf{e}$ is the all 1's vector. If we want to make the dependence of $h^\star(s)$ and $\lambda^\star$ on the underlying MDP $M$ explicit, we will write them as $h^\star(s; M)$ and $\lambda^\star(M)$ respectively.

In this paper, we give an algorithm REGAL, that achieves

$$\tilde{O}\left( \mathrm{sp}(h^\star(M)) S\sqrt{AT} \right)$$

regret with probability at least $1 - \delta$ when started in any state of a weakly communicating MDP $M$. Here $\mathrm{sp}(h)$ is the *span* of $h$ defined as,

$$\mathrm{sp}(h) := \max_{s \in \mathcal{S}} h(s) - \min_{s \in \mathcal{S}} h(s) .$$

The $\tilde{O}(\cdot)$ notation hides factors that are logarithmic in $S, A, T$ and $1/\delta$.

REGAL is based on the idea of *regularization* that has been successfully applied to obtain low regret algorithms in other settings involving decision making under uncertainty. The main idea is simple to describe. Using its experience so far, the agent can build a set $\mathcal{M}$ such that with high probability, the true MDP lies in $\mathcal{M}$. If the agent assumes that it is in the best of all possible worlds, it would choose an MDP $M' \in \mathcal{M}$ to maximize $\lambda^\star(M)$ and follow the optimal policy for $M'$. Instead, REGAL picks $M'$ to trade-off high gain for low span, by maximizing the following regularized objective,

$$\lambda^\star(M) - C\,\mathrm{sp}(h^\star(M)) .$$

### 1.1 RELATED WORK

The problem of simultaneous estimation and control of MDPs has been studied in the control theory [Kumar and Varaiya, 1986], operations research [Burnetas and Katehakis, 1997] and machine learning communities. In the machine learning literature, finite time bounds for undiscounted MDPs were pioneered by Kearns and Singh [2002]. Their seminal work spawned a long thread of research [Brafman and Tennenholtz, 2002, Kakade, 2003, Strehl and Littman, 2005, Strehl et al., 2006, Auer and Ortner, 2007, Tewari and Bartlett, 2008, Auer et al., 2009a]. These efforts improved the $S$ and $A$ dependence of their bounds, investigated lower bounds and explored other ways to study the exploration vs. exploitation trade-off.

The results of Auer and Ortner [2007] and Tewari and Bartlett [2008] applied to unichain and ergodic MDPs respectively. Recently, Auer et al. [2009a] gave an algorithm for communicating MDPs whose expected regret after $T$ steps is, with high probability,

$$\tilde{O}(D(M)S\sqrt{AT}) .$$

Here, $D(M)$ is the *diameter* of the MDP $M$ defined as

$$D(M) := \max_{s_1 \neq s_2} \min_\pi T^\pi_{s_1 \to s_2} ,$$

where $T^\pi_{s_1 \to s_2}$ is the expected number of steps it takes policy $\pi$ to get to $s_2$ from $s_1$. By definition, any communicating MDP has finite diameter. Let $\mathcal{O}$ denote the set of average-reward optimal policies. Previous work had considered various notions of "diameter", such as,

$$D_{\mathrm{worst}}(M) := \max_\pi \max_{s_1 \neq s_2} T^\pi_{s_1 \to s_2}$$

$$D_{\mathrm{opt}}(M) := \min_{\pi \in \mathcal{O}} \max_{s_1 \neq s_2} T^\pi_{s_1 \to s_2} .$$

Clearly $D \leq D_{\mathrm{opt}} \leq D_{\mathrm{worst}}$, and so, everything else being the same, an upper bound on regret in terms of $D$ is stronger.

We propose another diameter that we call the *one-way diameter*,

$$D_{\mathrm{ow}}(M) := \max_s \min_\pi T^\pi_{s_1 \to \bar{s}} , \quad (2)$$

where $\bar{s} = \mathrm{argmax}_s\ h^\star(s)$. We study the relationship between $\mathrm{sp}(h^\star)$, $D_{\mathrm{ow}}$ and $D$ in Section 4 below and prove that $\mathrm{sp}(h^\star) \leq D_{\mathrm{ow}} \leq D$ whenever the rewards are in $[0, 1]$. So, we not only extend the result of Auer et al. [2009a] to weakly communicating MDPs but also replace $D$ by a smaller quantity, $\mathrm{sp}(h^\star)$.

## 2 PRELIMINARIES

We make the simplifying assumption that the rewards $r(s,a) \in [0,1]$ are known and only the transition probabilities $P_{s,a}$ are unknown. This is not a restrictive assumption as the rewards can also be estimated at the cost of increasing the regret bound by some constant factor. The heart of the problem lies in not knowing the transition probabilities.

Recall the definition of the dynamic programming operator $\mathcal{T}$,

$$(\mathcal{T}V)(s) := \max_{a \in \mathcal{A}} \left( r(s,a) + P_{s,a}^\top V \right) .$$

Let $V^n(s)$ denote the maximum (over *all* policies, stationary or non-stationary) possible expected reward obtainable in $n$ steps starting from $s$. This can be computed via the simple recursion,

$$V^0(s) = 0$$
$$V^{n+1} = \mathcal{T}V^n .$$

Note that the optimality equations (1) can be written succinctly in terms of the operator $\mathcal{T}$,

$$h^\star + \lambda^\star \mathbf{e} = \mathcal{T}h^\star . \qquad (3)$$

## 3 REGAL

In this section, we present REGAL, an algorithm (see Algorithm 1) inspired by the UCRL2 algorithm of Auer et al. [2009a]. Before we describe the algorithm, let us set up some notation. Let $N(s,a,s';t)$ be the number of times the state-action-state triple $(s,a,s')$ has been visited up to time $t$. Let $N(s,a;t)$ be defined similarly. Then, an estimate of the state transition probability at time $t$ is

$$\hat{P}_{s,a}^t(s') = \frac{N(s,a,s';t)}{\max\{N(s,a;t),1\}} , \qquad (4)$$

Using these estimates, we can define a set $\mathcal{M}(t)$ of MDPs such that the true MDP lies in $\mathcal{M}^k$ with high probability. The set $\mathcal{M}(t)$ consists of all those MDPs whose transition probabilities satisfy,

$$\left\| P_{s,a} - \hat{P}_{s,a}^t \right\|_1 \le \sqrt{\frac{12S \log(2At/\delta)}{\max\{N(s,a,t),1\}}} . \qquad (5)$$

Like UCRL2, REGAL works in episodes. Let $t_k$ denote the start of episode $k$. We will abbreviate $N(s,a,s';t_k)$ and $N(s,a;t_k)$ to $N_k(s,a,s')$ and $N_k(s,a)$ respectively. Also, let $v_k(s,a) = N_{k+1}(s,a) - N_k(s,a)$ be the number of times $(s,a)$ is visited during episode $k$.

For choosing a policy to follow in episode $k$, REGAL first finds an MDP $M^k \in \mathcal{M}^k := \mathcal{M}(t_k)$ that maximizes a "regularized" average optimal reward,

$$\lambda^\star(M) - C_k \operatorname{sp}(h^\star(M)) . \qquad (6)$$

Here $C_k$ is a regularization parameter that is to be set appropriately at the start of episode $k$. REGAL then

---

**Algorithm 1** REGularization based Regret Minimizing ALgorithm (REGAL)
　**for** episodes $k = 1, 2, \ldots,$ **do**
　　$t_k \leftarrow$ current time
　　$\mathcal{M}^k$ is the set of MDPs whose transition function satisfies (5) with $t = t_k$
　　Choose $M^k \in \mathcal{M}^k$ to maximize the following criterion over $\mathcal{M}_k$,

$$\lambda^\star(M) - C_k \operatorname{sp}(h^\star(M)) .$$

　　$\pi^k \leftarrow$ average reward optimal policy for $M^k$
　　Follow $\pi^k$ until some $s, a$ pair gets visited $N_k(s,a)$ times
　**end for**

---

follows the average reward optimal policy $\pi^k$ for $M^k$ until some state-action pair is visited $N_k(s,a)$ times. When this happens, REGAL enters the next episode.

**Theorem 1.** *Suppose Algorithm 1 is run for $T$ steps in a weakly communicating MDP $M$ starting from some state $s_1$. Then, with probability at least $1 - \delta$, there exist choices for the regularization parameters $C_k$ such that,*

$$\Delta(s_1, T) = O\left(\operatorname{sp}(h^\star(M))S\sqrt{AT \log(AT/\delta)}\right) .$$

The proof of this theorem is given in Section 6. Unfortunately, as the proof reveals, the choice of the parameter $C_k$ in the theorem requires knowledge of the counts $v_k(s,a)$ *before* episode $k$ begins. Therefore, we cannot run REGAL with these choices of $C_k$. However, if an upper bound $H$ on the span $\operatorname{sp}(h^\star(M))$ of the true MDP $M$ is known, solving the constrained version of the optimization problem (6) gives Algorithm 2 that enjoys the following guarantee.

**Theorem 2.** *Suppose Algorithm 2 is run for $T$ steps in a weakly communicating MDP $M$ starting in some state $s_1$. Let the input parameter $H$ be such that $H \ge \operatorname{sp}(h^\star(M))$. Then, with probability at least $1 - \delta$,*

$$\Delta(s_1, T) = O\left(HS\sqrt{AT \log(AT/\delta)}\right) .$$

Finally, we also present a modification of REGAL that does not require knowledge of an upper bound on $\operatorname{sp}(h^*))$. The downside is that the regret guarantee we can prove becomes $\tilde{O}(\operatorname{sp}(h^\star)\sqrt{S^3AT})$. The modification, REGAL.D, that is given as Algorithm 3 uses the so called *doubling trick* to guess the length of an episode. If the guess turns out to be incorrect, the guess is doubled.

**Theorem 3.** *Suppose Algorithm 3 is run for $T$ steps in a weakly communicating MDP $M$ starting from*

some state $s_1$. Let the input parameter $c$ be chosen as
$$c = 2S\sqrt{12\log(2AT/\delta)} + \sqrt{2\log(1/\delta)} \ .$$
Then, with probability at least $1 - \delta$,
$$\Delta(s_1, T) = O\left(\mathrm{sp}(h^\star(M))\sqrt{S^3 AT \log^2(AT/\delta)}\right) \ .$$

---

**Algorithm 2** REGAL.C: Constrained Optimization version of REGAL

   Input parameter: $H$
   **for** episodes $k = 1, 2, \ldots$, **do**
      $t_k \leftarrow$ current time
      $\mathcal{M}^k$ is the set of MDPs whose transition function satisfies (5) with $t = t_k$
      Choose $M^k \in \mathcal{M}^k$ by solving the following optimization over $M \in \mathcal{M}_k$,
$$\max\ \lambda^\star(M) \quad \text{subject to} \quad \mathrm{sp}(h^\star(M)) \le H \ .$$
      $\pi^k \leftarrow$ average reward optimal policy for $M^k$
      Follow $\pi^k$ until some $s, a$ pair gets visited $N_k(s, a)$ times
   **end for**

---

**Algorithm 3** REGAL.D: REGAL with doubling trick

   Input parameter: $c$
   **for** episodes $k = 1, 2, \ldots$, **do**
      $t_k \leftarrow$ current time
      $\mathcal{M}^k$ is the set of MDPs whose transition function satisfies (5) with $t = t_k$
      $j \leftarrow 1$
      **while** every $(s, a)$ has been visited less than $N_k(s, a)$ times in this episode **do**
         $\ell_{k,j} \leftarrow 2^j$
         Choose $M^{k,j} \in \mathcal{M}^k$ to maximize the following criterion over $\mathcal{M}_k$,
$$\lambda^\star(M) - \frac{c}{\sqrt{\ell_{k,j}}}\mathrm{sp}(h^\star(M)) \ .$$
         $\pi^{k,j} \leftarrow$ average reward optimal policy for $M^{k,j}$
         Follow $\pi^{k,j}$ until:
         $\ell_{k,j}$ time steps have elapsed OR some $s, a$ pair gets visited $N_k(s, a)$ times
      **end while**
   **end for**

---

## 4 SPAN AND DIAMETER

**Theorem 4.** *In any weakly communicating MDP, for any states $s_1, s_2$, and any policy $\pi$,*
$$h^\star(s_2) - h^\star(s_1) \le \lambda^\star T^\pi_{s_1 \to s_2}$$

*Proof.* Let us first prove the theorem for all aperiodic weakly communicating MDPs. In such MDPs, value iteration is known to converge [Puterman, 1994]. That is, if we define a sequence of vectors using $v^{n+1} = \mathcal{T}v^n$ starting from an arbitrary $v^0$, then
$$\lim_{n \to \infty} v^n(s_1) - v^n(s_2) = h^\star(s_1) - h^\star(s_2) \ ,$$
for any $s_1, s_2$. If we choose $v^0 = \mathbf{0}$, then $v^n = V^n$ and hence we have
$$\lim_{n \to \infty} V^n(s_2) - V^n(s_1) = h^\star(s_2) - h^\star(s_1) \ ,$$
for any $s_1, s_2$. Let $\pi$ be a stationary policy. Consider the following $n$-step non-stationary policy. Follow $\pi$ starting from $s_1$ and wait until $s_2$ is reached. Suppose this happens in $\tau$ steps. Then follow the optimal $(n - \tau)$-step policy. Note that $\tau$ is a random variable and, by definition,
$$\mathbb{E}[\tau] = T^\pi_{s_1 \to s_2} \ .$$
The expected reward obtained in this way is at least $\mathbb{E}[V^{n-\tau}(s_2)]$. This has to be less than the $n$-step optimal reward. That is,
$$V^n(s_1) \ge \mathbb{E}[V^{n-\tau}(s_2)] \ .$$
Thus, we have
$$\begin{aligned}
h^\star(s_2) - h^\star(s_1) &= \lim_{n \to \infty} V^n(s_2) - V^n(s_1) \\
&\le \lim_{n \to \infty} V^n(s_2) - \mathbb{E}[V^{n-\tau}(s_2)] \\
&= \lim_{n \to \infty} \mathbb{E}[V^n(s_2) - V^{n-\tau}(s_2)] \\
&= \mathbb{E}[\lim_{n \to \infty} V^n(s_2) - V^{n-\tau}(s_2)] \\
&= \mathbb{E}[\lambda^\star \tau] \\
&= \lambda^\star T^\pi_{s_1 \to s_2} \ .
\end{aligned}$$

Now, if the MDP $M$ is periodic apply the following *aperiodicity transform* [Puterman, 1994] to get a new MDP $\tilde{M}$
$$\begin{aligned}
\tilde{r}(s, a) &= \theta r(s, a) \\
\tilde{P}_{s,a} &= (1 - \theta)\mathbf{e}_s + \theta P_{s,a} \ ,
\end{aligned}$$
where $\theta \in (0, 1)$. Note that $\tilde{M}$ is also weakly communicating. Let $\tilde{h}^\star, \tilde{\lambda}^\star$ and $\tilde{T}^\pi_{s_1 \to s_2}$ denote the quantities associated with $\tilde{M}$. It is easily verified that these are related to the corresponding quantities for $M$ as follows,
$$\begin{aligned}
\tilde{h}^\star &= h^\star \\
\tilde{\lambda}^\star &= \theta \lambda^\star \\
\tilde{T}^\pi_{s_1 \to s_2} &= \frac{T^\pi_{s_1 \to s_2}}{\theta} \ .
\end{aligned}$$

Using these relations and the fact that we have proved the result for $\tilde{M}$ gives us,

$$\begin{aligned}
h^\star(s_2) - h^\star(s_1) &= \tilde{h}^\star(s_2) - \tilde{h}^\star(s_1) \\
&\leq \tilde{\lambda}^\star \tilde{T}^\pi_{s_1 \to s_2} \\
&= \theta \lambda^\star \frac{T^\pi_{s_1 \to s_2}}{\theta} \\
&= \lambda^\star T^\pi_{s_1 \to s_2} .
\end{aligned}$$

Thus, we have proved the result for all weakly communicating MDPs. □

We can now derive the fact that $\text{sp}(h^\star) \leq D_{\text{ow}}$.

**Corollary 5.** *For any weakly communicating MDP $M$ with reward in $[0,1]$, the one-way diameter is an upper bound on the span of the optimal bias vector,*

$$\text{sp}(h^\star(M)) \leq D_{\text{ow}}(M) \leq D(M) .$$

*Moreover, both inequalities above can be arbitrarily loose.*

*Proof.* The inequalities follow immediately from Theorem 4. To show that they can be loose, consider a two-state two-action MDP $M$ such that

$$\begin{aligned}
P_{1,1} &= (1, 0)^\top , & P_{1,2} &= (1-\epsilon, \epsilon)^\top , \\
P_{2,1} &= (0, 1)^\top , & P_{2,2} &= (0, 1)^\top ,
\end{aligned}$$

and $r(1, a) = 1 - \alpha$, $r(2, a) = 1$ for all $a$. It is easy to verify that $\lambda^\star = 1$ and $h^\star = (-\alpha/\epsilon, 0)^\top$ satisfy the optimality equations (3). Thus, $\text{sp}(h^\star) = \alpha/\epsilon$ and $\bar{s} = 2$. Therefore, $D_{\text{ow}} = 1/\epsilon$. Moreover, $D = \infty$. Thus, both inequalities in the corollary can be arbitrarily loose. □

## 5 LOWER BOUND

Using Corollary 5, the bound of Theorem 1 can be written as $\tilde{O}\left(D_{\text{ow}}(M) S \sqrt{AT}\right)$. In this section, we provide a lower bound that matches the upper bound except for the dependence on $S$.

Using the results in Section 4, it is possible to show that the algorithms of Burnetas and Katehakis [1997] and Tewari and Bartlett [2008] enjoy a regret bound of $\tilde{O}(D_{\text{ow}}(M) \sqrt{SAT})$ for *ergodic* MDPs. We therefore conjecture that the lower bound is tight.

**Theorem 6.** *There exists a universal constant $c_0$ such that for any $S, A, d_{\text{ow}}$ and any algorithm $\mathcal{G}$, there exists an MDP $M$ with $D_{\text{ow}}(M) \leq d_{\text{ow}}$ such that for any $T > SA$ and any $s$, we have*

$$\Delta^{\mathcal{G}}(s, T) \geq c_0 d_{\text{ow}} \sqrt{SAT} .$$

*Proof Sketch.* Due to lack of space, we only outline the argument. We modify the MDP used by Auer et al. [2009a] to prove their lower bound. They build a large MDP with $S$ states, $A$ actions by putting together $S/2$ copies of an MDP with 2 states, $A$ actions. For our lower bound, we modify the construction of their 2-state MDP as follows. For all $a \in \mathcal{A}$, $P_{1,a} = (1-\alpha, \alpha)^\top$, $r(1, a) = 0$ and $r(2, a) = 1$. Thus, state 2 is the only rewarding state. For all but a single action $a^\star \in \mathcal{A}$, $P_{2,a} = (\delta, 1-\delta)^\top$ and for $a^\star$, set $P_{2,a^\star} = (\delta - \epsilon, 1 - \delta + \epsilon)$. We will choose $\alpha, \delta$ and $\epsilon$ later such that $\epsilon < \delta \ll \alpha$.

The idea is that taking $a^\star$ in state 2 increases the probability of staying in that rewarding state slightly. So, the agent needs to figure out which action is $a^\star$. To do this, each action needs to be probed at least $c\delta/\epsilon^2$ times. The difference in the average reward of the policy that takes action $a^\star$ in state 2 and the one that does not is

$$\frac{\alpha}{\alpha + \delta - \epsilon} - \frac{\alpha}{\alpha + \delta} > \frac{\epsilon}{4\alpha}$$

provided $\delta < \alpha$. Now connect the 1 states of the $S/2$ copies of the MDP using $A$ additional deterministic actions per state in an $A$-ary tree. Suppose only one of the copies has the good action $a^\star$ in it. Now, we need to probe at least $cSA\delta/\epsilon^2$ times. Choosing $\delta \ll \alpha$ means that most of the time, we are in state 2 and number of probes is a good approximation to the time elapsed. Thus, in probing for $a^\star$ we incur at least

$$\frac{cSA\delta}{\epsilon^2} \frac{\epsilon}{4\alpha} = \frac{cSA\delta}{4\alpha\epsilon}$$

regret. Now set $\epsilon = \delta\sqrt{SA/T}$ and $\delta = \alpha^2$. If $T > SA$, then $\epsilon < \delta$ and regret is at least

$$\frac{c\sqrt{SAT}}{4\alpha} .$$

Finally choose $\alpha = 1/d_{\text{ow}}$. Noting that $D_{\text{ow}}$ for this MDP is $1/\alpha$, proves the theorem (note that the MDP we have constructed has diameter $D = D_{\text{ow}}^2$). □

## 6 ANALYSIS

In this section we will prove Theorems 1, 2 and 3. We will first set up notation and recall key lemmas from the technical report of Auer et al. [2009b] that will be useful for proving all three theorems.

Let $M$ denote the true (but unknown) underlying MDP. The quantities $\lambda^\star$ and $h^\star$ will be for this MDP throughout this section. Let $\ell_k = \sum_{s,a} v_k(s, a)$ be the length of episode $k$. Let $\lambda_k, h_k^\star$ denote $\lambda(M^k), h^\star(M^k)$ respectively. The transition matrices of $\pi^k$ in $M^k$ and $M$ will be denoted by $\tilde{\mathbf{P}}_k$ and $\mathbf{P}_k$ respectively. Further, assume that $h^\star$ and $h_k^\star$ have been normalized to have their minimum component equal to 0, so that $\mathrm{sp}(h^\star) = \|h^\star\|_\infty$ and $\mathrm{sp}(h_k^\star) = \|h_k^\star\|_\infty$.

Let $\mathbf{v}_k, \mathbf{r}_k \in \mathbb{R}^S$ denote vectors such that
$$\mathbf{v}_k(s) = v_k(s, \pi^k(s)) \;,$$
$$\mathbf{r}_k(s) = r(s, \pi^k(s)) \;.$$

Note that, by definition of $\pi^k$, we have,
$$h_k^\star + \lambda_k^\star \mathbf{e} = \mathbf{r}_k + \tilde{\mathbf{P}}_k h_k^\star \;. \tag{7}$$

The following bound on the number $m$ of episodes until time $T$ was proved in Auer et al. [2009b].

**Lemma 7.** *The number $m$ of episodes of Algorithms 1,2 and 3 up to step $T \geq SA$ is upper bounded as,*
$$m \leq SA \log_2 \frac{8T}{SA} \;.$$

Let $\Delta_k$ be the regret incurred in episode $k$,
$$\Delta_k = \sum_{s,a} v_k(s,a)[\lambda^\star - r(s,a)] \;.$$

Note that the total regret equals,
$$\sum_{k \in G} \Delta_k + \sum_{k \in B} \Delta_k \;, \tag{8}$$

where $G = \{k \,:\, M \in \mathcal{M}^k\}$ and $B = \{1, \ldots, m\} - G$. The following result from Auer et al. [2009b] assures us that the contribution from "bad" episodes (in which the true MDP $M$ does not lie in $\mathcal{M}^k$) is small.

**Lemma 8.** *For Algorithms 1,2 and 3, with probability at least $1 - \delta$,*
$$\sum_{k \in B} \Delta_k \leq \sqrt{T} \;.$$

### 6.1 PROOF OF THEOREM 1

Consider an episode $k \in G$. Then, we have
$$\lambda_k^\star - C_k \mathrm{sp}(h_k^\star) \geq \lambda^\star - C_k \mathrm{sp}(h^\star) \;. \tag{9}$$

Therefore,
$$\Delta_k = \sum_{s,a} v_k(s,a)[\lambda^\star - r(s,a)]$$
$$\leq \sum_{s,a} v_k(s,a)[\lambda_k^\star - C_k \mathrm{sp}(h_k^\star) + C_k \mathrm{sp}(h^\star) - r(s,a)]$$
$$= \sum_{s,a} v_k(s,a)[\lambda_k^\star - r(s,a)]$$
$$\quad - C_k \sum_{s,a} v_k(s,a)[\mathrm{sp}(h_k^\star) - \mathrm{sp}(h^\star)]$$

$$= \mathbf{v}_k(\lambda_k^\star \mathbf{e} - \mathbf{r}_k) - C_k \sum_{s,a} v_k(s,a)[\mathrm{sp}(h_k^\star) - \mathrm{sp}(h^\star)]$$
$$= \mathbf{v}_k(\tilde{\mathbf{P}}_k - \mathbf{I})h_k^\star - C_k \sum_{s,a} v_k(s,a)[\mathrm{sp}(h_k^\star) - \mathrm{sp}(h^\star)]$$
[using (7)]
$$= \mathbf{v}_k(\tilde{\mathbf{P}}_k - \mathbf{P}_k + \mathbf{P}_k - \mathbf{I})h_k^\star$$
$$\quad - C_k \sum_{s,a} v_k(s,a)[\mathrm{sp}(h_k^\star) - \mathrm{sp}(h^\star)]$$
$$\leq \|\mathbf{v}_k(\tilde{\mathbf{P}}_k - \mathbf{P}_k)\|_1 \mathrm{sp}(h_k^\star) + \mathbf{v}_k(\mathbf{P}_k - \mathbf{I})h_k^\star$$
$$\quad - C_k \sum_{s,a} v_k(s,a)[\mathrm{sp}(h_k^\star) - \mathrm{sp}(h^\star)] \;. \tag{10}$$

**Lemma 9.** *With probability at least $1 - \delta$,*
$$\sum_{k \in G} \mathbf{v}_k(\mathbf{P}_k - \mathbf{I})h_k^\star$$
$$\leq \sum_{k \in G} \mathrm{sp}(h_k^\star)\sqrt{2\ell_k \log(1/\delta)}$$
$$\quad + (\mathrm{sp}(h^\star) + \max_{k \in G} \frac{1}{C_k})(m + \log(1/\delta)) \;.$$

*Proof.* Using Bernstein's inequality (see, for example, [Cesa-Bianchi and Lugosi, 2006, Lemma A.8]) instead of Hoeffding-Azuma in the argument of [Auer et al., 2009b, Section B.4], we get, with probability at least $1 - \delta$,
$$\sum_{k \in G} \mathbf{v}_k(\mathbf{P}_k - \mathbf{I})h_k^\star \leq \sqrt{2 \sum_{k \in G} \mathrm{sp}(h_k^\star)^2 \ell_k \log(1/\delta)}$$
$$\quad + \max_{k \in G} \mathrm{sp}(h_k^\star)(m + \log(1/\delta)) \;.$$

Equation (9) gives $\mathrm{sp}(h_k^\star) \leq \mathrm{sp}(h^\star) + 1/C_k$. Using this and sub-additivity of square-root gives the lemma. $\square$

If $k \in G$ then $M^k, M \in \mathcal{M}^k$ and so we also have
$$\|\mathbf{v}_k(\tilde{\mathbf{P}}_k - \mathbf{P}_k)\|_1 \leq 2 \sum_{s,a} v_k(s,a)\sqrt{\frac{12S \log(2AT/\delta)}{N_k(s,a)}} \tag{11}$$

Using this, Lemma 9 and plugging these into (10), we get
$$\sum_{k \in G} \Delta_k \leq \sum_{k \in G} \mathrm{sp}(h_k^\star) \left( 2 \sum_{s,a} v_k(s,a)\sqrt{\frac{12S \log(2AT/\delta)}{N_k(s,a)}} \right.$$
$$\left. + \sqrt{2\ell_k \log(1/\delta)} - C_k \sum_{s,a} v_k(s,a) \right)$$
$$\quad + \sum_{k \in G} C_k \ell_k \mathrm{sp}(h^\star)$$
$$\quad + (\mathrm{sp}(h^\star) + \max_{k \in G} \frac{1}{C_k})(m + \log(1/\delta)) \;.$$

Equation (9) in Auer et al. [2009b] gives,

$$\sum_k \sum_{s,a} \frac{v_k(s,a)}{N_k(s,a)} \leq \sqrt{8SAT} \ . \quad (12)$$

Therefore, choosing $C_k$ to satisfy,

$$C_k = \frac{2\sum_{s,a} v_k(s,a)\sqrt{\frac{12S \log \frac{2AT}{\delta}}{N_k(s,a)}} + \sqrt{2\ell_k \log \frac{1}{\delta}}}{\ell_k}$$

gives us

$$\sum_{k \in G} \Delta_k \leq \sum_{k \in G} C_k \ell_k \operatorname{sp}(h^\star)$$
$$+ (\operatorname{sp}(h^\star) + \sqrt{\frac{T}{2 \log(1/\delta)}}(m + \log(1/\delta))$$
$$= O\left(\operatorname{sp}(h^\star) S\sqrt{AT \log(AT/\delta)}\right)$$

Combining this with (8) and Lemma 8 finishes the proof of Theorem 1. □

## 6.2 PROOF OF THEOREM 2

Consider an episode $k \in G$. Then, we have $\lambda_k^\star \geq \lambda^\star$. Therefore,

$$\Delta_k = \sum_{s,a} v_k(s,a)[\lambda^\star - r(s,a)]$$
$$\leq \sum_{s,a} v_k(s,a)[\lambda_k^\star - r(s,a)]$$
$$= \mathbf{v}_k(\lambda_k^\star \mathbf{e} - \mathbf{r}_k) = \mathbf{v}_k(\tilde{\mathbf{P}}_k - \mathbf{I})h_k^\star$$
$$= \mathbf{v}_k(\tilde{\mathbf{P}}_k - \mathbf{P}_k + \mathbf{P}_k - \mathbf{I})h_k^\star$$
$$\leq \|\mathbf{v}_k(\tilde{\mathbf{P}}_k - \mathbf{P}_k)\|_1 \operatorname{sp}(h_k^\star) + \mathbf{v}_k(\mathbf{P}_k - \mathbf{I})h_k^\star \ . \quad (13)$$

The following lemma is proved like Lemma 9 (in this case, Hoeffding-Azuma suffices).

**Lemma 10.** *With probability at least $1 - \delta$,*

$$\sum_{k \in G} \mathbf{v}_k(\mathbf{P}_k - \mathbf{I})h_k^\star \leq H\left(\sqrt{2T \log(1/\delta)} + m\right) \ .$$

Equation (11) still holds if $k \in G$. Plugging it and Lemma 10 into (13), we get

$$\sum_{k \in G} \Delta_k \leq H\left(\sum_{k \in G} \sum_{s,a} 2v_k(s,a)\sqrt{\frac{12S \log(2AT/\delta)}{N_k(s,a)}}\right.$$
$$\left. + \sqrt{2T \log(1/\delta)} + m\right) \ .$$

Using (12) now gives,

$$\sum_{k \in G} \Delta_k \leq O(HS\sqrt{AT \log(AT/\delta)}) \ .$$

Combining this with (8) and Lemma 8 finishes the proof of Theorem 2. □

## 6.3 PROOF OF THEOREM 3

In this case, we have episodes consisting of several sub-episodes whose lengths increase in geometric progression. Let $v_{k,j}(s,a)$ be the number of times $(s,a)$ is visited during sub-episode $j$ of episode $k$. Thus, $\ell_{k,j} = \sum_{s,a} v_{k,j}(s,a)$. Let $\lambda_{k,j}, h_{k,j}^\star$ denote $\lambda(M^{k,j}), h^\star(M^{k,j})$ respectively. The transition matrices of $\pi^{k,j}$ in $M^{k,j}$ and $M$ will be denoted by $\tilde{\mathbf{P}}_{k,j}$ and $\mathbf{P}_{k,j}$ respectively. Let $\mathbf{v}_{k,j}, \mathbf{r}_{k,j}$ be defined accordingly.

Consider an episode $k \in G$. Then, as in the proof of Theorem 1, we have,

$$\lambda_{k,j}^\star - C_{k,j} \operatorname{sp}(h_{k,j}^\star) \geq \lambda^\star - C_{k,j} \operatorname{sp}(h^\star) \ ,$$

and therefore,

$$\Delta_{k,j}$$
$$:= \sum_{s,a} v_{k,j}(s,a)[\lambda^\star - r(s,a)]$$
$$\leq \sum_{s,a} v_{k,j}(s,a)[\lambda_{k,j}^\star - C_{k,j} \operatorname{sp}(h_{k,j}^\star)$$
$$\qquad + C_{k,j} \operatorname{sp}(h^\star) - r(s,a)]$$
$$= \mathbf{v}_{k,j}(\lambda_{k,j}^\star \mathbf{e} - \mathbf{r}_{k,j})$$
$$\quad - C_{k,j} \sum_{s,a} v_{k,j}(s,a)[\operatorname{sp}(h_{k,j}^\star) - \operatorname{sp}(h^\star)]$$
$$= \mathbf{v}_{k,j}(\tilde{\mathbf{P}}_{k,j} - \mathbf{I})h_{k,j}^\star$$
$$\quad - C_{k,j} \sum_{s,a} v_{k,j}(s,a)[\operatorname{sp}(h_{k,j}^\star) - \operatorname{sp}(h^\star)]$$
$$\leq \|\mathbf{v}_{k,j}(\tilde{\mathbf{P}}_{k,j} - \mathbf{P}_{k,j})\|_1 \operatorname{sp}(h_{k,j}^\star) + \mathbf{v}_{k,j}(\mathbf{P}_{k,j} - \mathbf{I})h_{k,j}^\star$$
$$\quad - C_{k,j} \sum_{s,a} v_{k,j}(s,a)[\operatorname{sp}(h_{k,j}^\star) - \operatorname{sp}(h^\star)] \ . \quad (14)$$

If $k \in G$ then $M^{k,j}, M \in \mathcal{M}^k$. Therefore,

$$\|\mathbf{v}_{k,j}(\tilde{\mathbf{P}}_{k,j} - \mathbf{P}_k)\|_1$$
$$\leq 2 \sum_{s,a} v_{k,j}(s,a)\sqrt{\frac{12S \log(2AT/\delta)}{N_k(s,a)}}$$
$$\leq 2\sqrt{12S \log(2AT/\delta)}\sqrt{\sum_{s,a} v_{k,j}(s,a)} \cdot \sqrt{\sum_{s,a} \frac{v_{k,j}(s,a)}{N_k(s,a)}}$$
$$\leq 2\sqrt{12S \log(2AT/\delta)}\sqrt{\ell_{k,j}} \cdot \sqrt{S}$$
$$\quad [\because v_{k,j}(s,a) \leq v_k(s,a) \leq N_k(s,a)]$$
$$= 2S\sqrt{12\ell_{k,j} \log(2AT/\delta)} \quad (15)$$

Note that this step differs slightly from its counterpart in the proof of Theorem 1. Here, we used Cauchy-Schwarz (for the second inequality above) to get a bound solely in terms of the length $\ell_{k,j}$ of the sub-episode. This allows us to deal with the problem of not knowing the visit counts $v_{k,j}(s,a)$ in advance.

The following lemma is proved exactly like Lemma 9.

**Lemma 11.** *With probability at least $1 - \delta$,*

$$\sum_{k \in G} \sum_j \mathbf{v}_{k,j}(\mathbf{P}_{k,j} - \mathbf{I})h^\star_{k,j}$$
$$\leq \sum_{k \in G} \sum_j \operatorname{sp}(h^\star_{k,j})\sqrt{2\ell_{k,j}\log(1/\delta)}$$
$$+ (\operatorname{sp}(h^\star) + \max_{k,j} \frac{1}{C_{k,j}})(m\log_2(T) + \log(1/\delta)) \ .$$

Combining (15) and Lemma 11 with (14), we have,

$$\sum_{k \in G} \Delta_k$$
$$= \sum_{k \in G} \sum_j \Delta_{k,j}$$
$$\leq \sum_{k \in G} \sum_j \operatorname{sp}(h^\star_{k,j}) \left( 2S\sqrt{12\ell_{k,j}\log(2AT/\delta)} \right.$$
$$\left. + \sqrt{2\ell_{k,j}\log(1/\delta)} - C_{k,j}\ell_{k,j} \right)$$
$$+ \operatorname{sp}(h^\star) \sum_{k \in G} \sum_j C_{k,j}\ell_{k,j}$$
$$+ (\operatorname{sp}(h^\star) + \max_{k,j} \frac{1}{C_{k,j}})(m\log_2(T) + \log(1/\delta)) \ .$$

Now, setting $C_{k,j} = c/\sqrt{\ell_{k,j}}$ for

$$c = 2S\sqrt{12\log(2AT/\delta)} + \sqrt{2\log(1/\delta)}$$

gives us,

$$\sum_{k \in G} \Delta_k$$
$$\leq \operatorname{sp}(h^\star) \sum_{k \in G} \sum_j c\sqrt{\ell_{k,j}} \quad (16)$$
$$+ (\operatorname{sp}(h^\star) + c\max_{k,j} \sqrt{\ell_{k,j}})(m\log_2(T) + \log(1/\delta)) \ .$$

Since $\ell_{k,j} = 2^j$, $\sum_j \sqrt{\ell_{k,j}} \leq 6\sqrt{\ell_k + 2}$. Therefore,

$$\sum_{k \in G} \sum_j \sqrt{\ell_{k,j}} \leq \sum_{k \in G} 6\sqrt{\ell_k + 2}$$
$$\leq 6\sqrt{m}\sqrt{\sum_k \ell_k + 2m}$$
$$= 6\sqrt{m}\sqrt{T + 2m} \ .$$

Substituting this into (16), gives

$$\sum_{k \in G} \Delta_k = O\left(\operatorname{sp}(h^\star)\sqrt{S^3 AT \log^2(AT/\delta)}\right) \ .$$

Combining this with (8) and Lemma 8 finishes the proof of Theorem 3. $\square$

**Acknowledgments**

We gratefully acknowledge the support of DARPA under award FA8750-05-2-0249.

**References**

Peter Auer and Ronald Ortner. Logarithmic online regret bounds for undiscounted reinforcement learning. In *Advances in Neural Information Processing Systems 19*, pages 49–56. MIT Press, 2007.

Peter Auer, Thomas Jaksch, and Ronald Ortner. Near-optimal regret bounds for reinforcement learning. In *Advances in Neural Information Processing Systems 21*, 2009a.

Peter Auer, Thomas Jaksch, and Ronald Ortner. Near-optimal regret bounds for reinforcement learning (full version), 2009b. URL: http://institute.unileoben.ac.at/infotech/publications/ucrl2.pdf.

Ronen I. Brafman and Moshe Tennenholtz. R-MAX – a general polynomial time algorithm for near-optimal reinforcement learning. *Journal of Machine Learning Research*, 3:213–231, 2002.

A. N. Burnetas and M. N. Katehakis. Optimal adaptive policies for Markov decision processes. *Mathematics of Operations Research*, 22(1):222–255, 1997.

Nicolò Cesa-Bianchi and Gábor Lugosi. *Prediction, Learning, and Games*. Cambridge University Press, 2006.

Sham Kakade. *On the Sample Complexity of Reinforcement Learning*. PhD thesis, Gatsby Computational Neuroscience Unit, University College London, 2003.

Michael Kearns and Satinder Singh. Near-optimal reinforcement learning in polynomial time. *Machine Learning*, 49:209–232, 2002.

P. R. Kumar and P. P. Varaiya. *Stochastic systems: Estimation, identification, and adaptive control*. Prentice Hall, 1986.

Martin L. Puterman. *Markov Decision Processes: Discrete Stochastic Dynamic Programming*. John Wiley & Sons, 1994.

Alexander L. Strehl and Michael Littman. A theoretical analysis of model-based interval estimation. In *Proceedings of the Twenty-Second International Conference on Machine Learning*, pages 857–864. ACM Press, 2005.

Alexander L. Strehl, Lihong Li, Eric Wiewiora, John Langford, and Michael L. Littman. PAC model-free reinforcement learning. In *Proceedings of the Twenty-Third International Conference on Machine Learning*, 2006.

Ambuj Tewari and Peter L. Bartlett. Optimistic linear programming gives logarithmic regret for irreducible MDPs. In *Advances in Neural Information Processing Systems 20*, pages 1505–1512. MIT Press, 2008.